\documentclass[10pt,twocolumn,letterpaper]{article}
\usepackage[pagenumbers]{cvpr}
\usepackage{amsthm}
\usepackage{tabularx}
\usepackage{array}
\usepackage{multirow}
\usepackage{microtype}
\usepackage{algorithm}
\usepackage{algpseudocode}
\definecolor{cvprblue}{rgb}{0.21,0.49,0.74}
\definecolor{objectiveblue}{RGB}{40,120,181}
\definecolor{objectiveorange}{RGB}{213,94,0}

\usepackage[pagebackref,breaklinks,colorlinks,allcolors=cvprblue]{hyperref}
\newtheorem{proposition}{Proposition}
\newtheorem{observation}{Observation}
\hypersetup{pdfauthor={Pu Cao, Min Zhou, Ying Shu, Tiezheng Ge, Qing Song, Lu Yang},pdftitle={When Anchors Shift the Target: Advantage-Gated Adaptation for Diffusion Reinforcement Learning},pdfsubject={arXiv preprint}}

\usepackage{xcolor}
\usepackage{colortbl}
\definecolor{taskrewardblue}{RGB}{229,243,255}
\definecolor{offsetgray}{RGB}{242,242,242}
\definecolor{ablatedgray}{RGB}{110,110,110}
\newcommand{\rewardcell}[1]{\cellcolor{taskrewardblue}#1}

\usepackage{placeins}
\title{When Anchors Shift the Target:\\Advantage-Gated Adaptation for Diffusion Reinforcement Learning}
\author{
Pu Cao$^{1}$ \quad
Min Zhou$^{2}$ \quad
Ying Shu$^{2}$ \quad
Tiezheng Ge$^{2}$ \quad
Qing Song$^{1}$ \quad
Lu Yang$^{1}$\thanks{Corresponding author.}\\[0.5em]
{\normalsize $^{1}$Beijing University of Posts and Telecommunications \quad $^{2}$Alibaba Group}\\[0.25em]
{\small\texttt{\{caopu, priv, soeaver\}@bupt.edu.cn}}
}
\begin{document}
\raggedbottom
\maketitle
\begin{abstract}
Forward-process reinforcement learning adapts diffusion and flow-matching models by re-noising rewarded generations into regression targets. Because negative advantages induce concave updates with unbounded repulsive gradients, existing methods stabilize training using cached predictions from a moving policy copy, yet these anchors also alter what positive feedback teaches the model to fit. We formalize this conflict as \emph{anchor coupling}: after separating frozen-reference regularizers, affine moving-anchor rules cannot eliminate negative curvature while preserving every positive regression target. We propose \emph{Advantage-Gated Adaptation (AGA)} to decouple these roles. Positive samples retain standard target fitting, whose self-limiting update vanishes at the observed target independently of the moving anchor. Negative samples instead use the first-order tangent of the concave objective at the moving-policy prediction, preserving the local repulsion direction with a prediction-independent gradient. Within the quadratic-anchor family, these requirements uniquely determine the parameter-free gate $\omega(a)=\max(-a,0)$. Because the moving copy provides local linearization rather than retention, AGA assigns long-horizon behavior preservation to a frozen pretrained reference regularized at its operating classifier-free guidance scale, interpreted through an implicit-reward lens. Across compositional generation, text rendering, and preference alignment, our experiments show that displacing positive targets degrades visual quality and heightens sensitivity to policy-copy schedules, whereas AGA consistently improves the reward--quality trade-off.
\end{abstract}

\section{Introduction}
\label{sec:introduction}

Diffusion and flow-matching models learn generative dynamics through velocity or noise prediction objectives \cite{ho2020denoising,lipman2023flow}, powering modern latent text-to-image synthesis \cite{rombach2022high,esser2024scaling}. Online reinforcement learning (RL) adapts these models to downstream goals such as compositional object placement, legible text rendering, and human preference alignment using scalar rewards evaluated on model-generated samples \cite{black2024training,fan2023dpok,liu2025flowgrpo}. Rather than backpropagating through multi-step sampling chains or estimating trajectory-level likelihood ratios, forward-process RL brings reward feedback directly into the pretraining interface: sample an image, evaluate its reward, re-noise it along the forward path, and update a velocity-matching objective at independent noised states \cite{xue2025advantage,zheng2026diffusionnft,bergmeister2026ram}.

Positive and negative advantages play fundamentally asymmetric roles in this local regression problem. A positive-advantage sample supplies a concrete velocity target to fit, whereas a negative-advantage sample indicates an undesirable generation to move away from. Direct signed matching applies a weighted squared error to both branches, but flipping the sign makes the negative branch concave: its repulsive gradient grows linearly with prediction error, driving self-amplifying runaway updates as the policy moves away from the penalized target. Recent methods therefore incorporate cached predictions from a moving policy copy as quadratic penalties or target corrections to restore local stability \cite{xue2025advantage,zheng2026diffusionnft,kveton2026advantageflow,fu2026flowawr,bergmeister2026ram}.

Yet stabilizing negative updates with a moving anchor can also alter what positive feedback teaches the model to fit. Separating frozen-reference regularizers, we show that widely used forward-process objectives share a unified local quadratic form whose anchor weight is affine in the signed advantage. Within this family, removing negative curvature inevitably shifts the stationary target of positive samples: stabilizing the negative branch pulls positive targets toward the moving prediction, rescales them by advantage magnitude, or eliminates a unique stationary target altogether. We call this conflict \emph{anchor coupling}. The key design question is therefore \emph{how to stabilize negative repulsion without shifting the regression targets supplied by positive feedback}.

We propose \emph{Advantage-Gated Adaptation (AGA)} to decouple these two roles. On positive samples, AGA retains standard target regression: advantage scales fitting strength while updates follow the live prediction residual, remaining self-limiting and vanishing at the observed target regardless of the moving anchor. On negative samples, AGA replaces the concave quadratic with its first-order tangent upper bound expanded at the moving-policy prediction, preserving the local repulsion direction while keeping the prediction-space gradient independent of the online prediction. Within the quadratic-anchor family, these two criteria uniquely determine the parameter-free gate $\omega(a)=\max(-a,0)$, under which the moving copy sets the negative expansion point without displacing positive targets.

Because this gated moving anchor provides zero-curvature linearization rather than quadratic retention, it does not constrain long-horizon policy drift. AGA therefore assigns global behavior preservation to a separate frozen pretrained reference. Recent forward-process RL methods often adopt CFG-free ($\gamma=1$) training for efficiency, with DiffusionNFT interpreting RL as internalizing guidance without explicit CFG~\cite{xue2025advantage,zheng2026diffusionnft}. Yet classifier-free guidance (CFG) \cite{ho2022classifierfree} still governs the effective sampling distribution of the pretrained model even with fixed weights, so $\gamma=1$ regularizes toward a weak unguided prior rather than the model's preferred guided behavior. We therefore regularize the composed guided velocity field at the operating guidance scale used for rollout and evaluation, and ground this reference choice in an endpoint implicit-reward view.

Experiments on SD3.5-Medium across compositional generation, text rendering, and preference alignment show improved task adaptation and reward--quality trade-offs, including gains under matched CFG-free ($\gamma=1$) training. Our contributions are:
\begin{itemize}
 \item We formalize \emph{anchor coupling}, proving that affine moving-anchor rules cannot eliminate negative curvature while preserving every positive regression target.
 \item We derive AGA's gate $\omega(a)=\max(-a,0)$ from positive target preservation and negative tangent linearization, and pair it with guided frozen-reference retention to separate local feedback from behavior preservation.
 \item Through controlled studies of feedback branches, policy-copy updates, and guidance scales, we show that preserving positive targets prevents quality collapse and sustains a superior reward--quality balance.
\end{itemize}

\section{Related Work}
\label{sec:related}

\paragraph{Reward-based visual post-training.}
Trajectory-based RL methods such as DDPO and DPOK optimize online rewards through multi-step denoising policies~\cite{black2024training,fan2023dpok,schulman2017proximal}, a paradigm extended to flow and diffusion generators by Flow-GRPO and DanceGRPO~\cite{liu2025flowgrpo,xue2025dancegrpo}. Alternative approaches backpropagate differentiable rewards through sampling trajectories~\cite{clark2024draft} or align models offline from pairwise and temporal preferences~\cite{wallace2024diffusiondpo,yang2024dense}. Because aggressive reward maximization frequently degrades visual realism and diversity through reward over-optimization, recent work tracks general image quality on held-out prompts alongside task rewards~\cite{kim2024confidence,li2026diffusionopd}. Focusing on online forward-process RL, we study how anchor design in this velocity-matching interface governs the trade-off between task reward and held-out image quality.

\paragraph{Matching objectives and policy anchoring.}
Forward-process RL methods differ in how they incorporate cached moving-policy predictions to stabilize negative-advantage updates. AWM combines advantage-weighted matching with an EMA velocity penalty~\cite{xue2025advantage}, while AdvantageFlow sets promptwise anchor weights to enforce local strong convexity and reduce gradient variance~\cite{kveton2026advantageflow}. DiffusionNFT constructs implicit positive and negative velocity predictions to form a convex regression loss~\cite{zheng2026diffusionnft}, FlowAWR regresses toward shifted targets derived from a reward-tilted velocity field~\cite{fu2026flowawr}, and RAM obtains reward-corrected targets from optimal control~\cite{bergmeister2026ram}, with RVM providing a unified velocity-matching parameterization~\cite{choi2026rvm}. Unlike affine anchor rules across this family, which couple negative-curvature stabilization to positive-target shifts, AGA decouples the two feedback signs.

\paragraph{Guidance and pretrained references.}
Classifier-free guidance (CFG) trades diversity for fidelity by extrapolating conditional and unconditional predictions~\cite{dhariwal2021diffusion,ho2022classifierfree}. Diff-Instruct++ connects CFG distillation to implicit rewards~\cite{luo2024diffinstructpp}, while Decoupled DMD and Reinforcement Learning Guidance study guidance in distribution matching and alignment~\cite{liu2025decoupleddmd,jin2025rlg}. While recent forward-process RL methods use CFG-free ($\gamma=1$) training for efficiency~\cite{xue2025advantage,zheng2026diffusionnft}, frozen references preserve pretrained capabilities during RL~\cite{fan2023dpok,kveton2026advantageflow}, including guided reference matching in CRD and RVM~\cite{zhu2026crd,choi2026rvm}. Building on these guided regularizers, we study when CFG-free anchoring degrades visual quality, ground reference-scale selection in an implicit-reward criterion, and regularize the composed guided field throughout adaptation.

\section{Advantage-Gated Adaptation}
\label{sec:method}

We first analyze how moving-policy anchors couple negative-update stabilization to positive target placement (Section~\ref{sec:policy}). We then derive an advantage-gated objective that decouples these roles (Section~\ref{sec:aga}) and pair it with guided frozen-reference retention (Section~\ref{sec:prior}). Section~\ref{sec:overview} summarizes the complete training procedure.

\subsection{Policy Anchoring in Diffusion RL}
\label{sec:policy}

\paragraph{Learning rewards through matching.}
For a prompt $c$, a behavior policy generates an image $x_0$ and receives reward $r(x_0,c)$. Forward-process RL methods train on noised states of these rewarded images \cite{xue2025advantage,zheng2026diffusionnft}. With $\epsilon\sim\mathcal N(0,I)$, define
\begin{equation}
 x_t=(1-t)x_0+t\epsilon,\qquad u=\epsilon-x_0,
 \quad t\in[0,1].
 \label{eq:path}
\end{equation}
Let $v=v_\theta^c(x_t,t,c)$ be the conditional velocity prediction. Within each prompt group, reward-derived advantages define a detached signed weight $a$ ($a\in[-1,1]$ in AGA), giving the sample matching objective
\begin{equation}
 \ell_a(v)=\frac a2\|v-u\|^2,
 \qquad -\nabla_v\ell_a=a(u-v),
 \label{eq:signed}
\end{equation}
aggregated across timesteps with nonnegative weights. Our comparison below expresses each method in terms of its own effective signed matching weight $a$.

\begin{figure}[!t]
 \centering
 \includegraphics[width=\columnwidth]{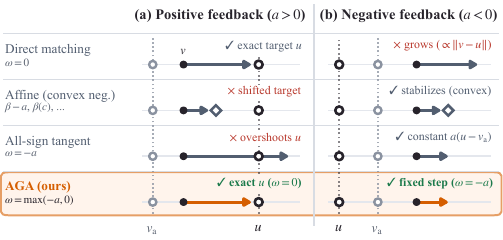}
 \caption{\textbf{Anchor rules across feedback signs.} (a) For $a>0$, AGA hits $u$, while affine rules shift the target and all-sign tangent overshoots. (b) For $a<0$, AGA fixes the step via $v_{\rm a}$ instead of growing with $\|v-u\|$.}
 \label{fig:method-roles}
\end{figure}

\paragraph{Asymmetric roles of positive and negative feedback.}
For $a>0$, Eq.~\eqref{eq:signed} is strictly convex with a unique minimum at $u$: the update $a(u-v)$ is self-limiting, pointing toward $u$ and vanishing as $v\to u$, so advantage scales fitting strength without shifting the target. For $a<0$, the quadratic is concave, and the repulsive update points away from $u$ with magnitude $|a|\|v-u\|$, which is self-amplifying and grows without bound as $v$ moves away from $u$. Unlike positive feedback, negative feedback specifies an outcome to avoid rather than an alternative target to fit.

\paragraph{Moving anchors and anchor coupling.}
Because the frozen-reference regularizer is advantage-independent and shared across post-training methods to anchor global behavior~\cite{fan2023dpok,xue2025advantage,choi2026rvm}, factoring it out isolates how each method uses a cached moving-policy prediction $v_{\rm a}=\operatorname{sg}[v_{\theta_{\rm a}}^c(x_t,t,c)]$ inside penalties or modified targets \cite{xue2025advantage,zheng2026diffusionnft,kveton2026advantageflow,fu2026flowawr,bergmeister2026ram}, where $\operatorname{sg}$ denotes stop-gradient. Define $D=\tfrac12\|v-u\|^2$ and $K=\tfrac12\|v-v_{\rm a}\|^2$. Separating frozen-reference terms, these reward-matching objectives take the local quadratic form
\begin{equation}
 \ell_{a,\omega}(v)=aD+\omega(a)K.
 \label{eq:anchored}
\end{equation}
We examine whether $\omega(a)$ can remove negative curvature without shifting positive targets. With sample coefficients held fixed, the prediction Hessian and stationary point are
\begin{equation}
 \begin{aligned}
  \nabla_v^2\ell_{a,\omega}&=(a+\omega)I,\\
  v^\star&=\frac{au+\omega v_{\rm a}}{a+\omega},\qquad a+\omega>0.
 \end{aligned}
 \label{eq:anchor-properties}
\end{equation}
Set $b=|a|>0$ and $\Delta=u-v_{\rm a}$. For negative samples ($a=-b$), the curvature coefficient is $\omega(-b)-b$. For positive samples ($a=b$), the stationary target is $v^\star=v_{\rm a}+s(b)\Delta$, with positive target scale
\begin{equation}
 s(b)=\frac{b}{b+\omega(b)},\qquad b+\omega(b)>0.
 \label{eq:positive-reach}
\end{equation}
Recovering the observed target $u$ for every anchor $v_{\rm a}$ requires $s(b)=1$ ($\omega(b)=0$).

\begin{table}[!t]
 \centering
 \caption{\textbf{Effects of moving anchors on fitting.} AGA removes negative curvature while preserving the positive target.}
 \label{tab:policy-anchors}
 \begingroup
\fontsize{7.5}{8.5}\selectfont
\setlength{\tabcolsep}{1.4pt}
\renewcommand{\arraystretch}{1.15}
\begin{tabular*}{\columnwidth}{@{\extracolsep{\fill}}lccc@{}}
\toprule
Method & $\omega(a)$ & Neg. curvature & Pos. scale $s(b)$\\
\midrule
Direct matching & $0$ & $-b$ (concave) & $1$ (exact)\\
\midrule
AWM & $\beta$ & $\beta-b$ & $\frac{b}{b+\beta}$ (contracted)\\
AdvantageFlow & $\beta(c)$ & $\geq\gamma_0$ (convex) & $\frac{b}{b+\beta(c)}$ (contracted)\\
DiffusionNFT & $\beta-a$ & $\beta$ (convex) & $b/\beta$ (rescaled)\\
FlowAWR & $1-a$ & $1$ (convex) & $b$ (rescaled)\\
RAM & $-a$ & $0$ (linear) & No unique target\\
\midrule
\textbf{AGA} & $\max(-a,0)$ & $0$ (linear) & $\mathbf{1}$ (exact)\\
\bottomrule
\end{tabular*}
\endgroup

\end{table}

Table~\ref{tab:policy-anchors} compares these two effects across methods (full derivations are in Appendix~A.2). Existing objectives arrive at Eq.~\eqref{eq:anchored} through two main mechanisms: AWM~\cite{xue2025advantage} and AdvantageFlow~\cite{kveton2026advantageflow} add an explicit velocity penalty $\omega K$ with $\omega>0$, whereas DiffusionNFT~\cite{zheng2026diffusionnft}, FlowAWR~\cite{fu2026flowawr}, and RAM~\cite{bergmeister2026ram} construct interpolated predictions or shifted targets along $\Delta=u-v_{\rm a}$, which expand algebraically into $aD+(\beta-a)K$ up to $v$-independent constants (with $\beta=1$ for FlowAWR and $\beta=0$ for RAM). Because all of these rules parameterize $\omega(a)$ as an affine function of $a$, the same coefficients govern both branches, producing an unavoidable conflict.

\begin{observation}[Anchor coupling]
\label{obs:anchor-coupling}
For promptwise constants $(\alpha,\kappa)$, an affine rule $\omega(a)=\alpha+\kappa a$ that removes negative curvature cannot preserve $u$ as the unique fitting point across any positive-advantage interval when $v_{\rm a}\neq u$.
\end{observation}
Maintaining $s(b)\equiv1$ requires $\omega(b)=0$ throughout that interval; an affine function with this property vanishes identically, leaving negative curvature $-b$. Consequently, affine rules distort positive fitting in three distinct ways (Table~\ref{tab:policy-anchors}). First, constant or promptwise positive anchors $\omega(b)>0$ \emph{contract} the stationary target toward $v_{\rm a}$ ($s(b)<1$ in AWM, and $s(b)<1/2$ under AdvantageFlow's $\beta(c)=\max_i|a_i|+\gamma_0$), so strengthening negative-branch convexity directly dampens progress toward positive targets. Second, rules of the form $\omega(a)=\beta-a$ \emph{rescale} the target step to $s(b)=b/\beta$ (DiffusionNFT and FlowAWR), causing samples with $b<\beta$ to fall short of $u$ and samples with $b>\beta$ to extrapolate beyond $u$. Third, setting $\omega(a)=-a$ (RAM's moving-anchor component) makes both branches affine in $v$, eliminating a unique stationary target on positive samples altogether.

\subsection{Advantage-Gated Anchoring}
\label{sec:aga}

To decouple these effects, we impose separate requirements on the two signs of feedback. Positive samples must retain $u$ as their unique stationary target for every $v_{\rm a}$ so updates remain self-limiting. Negative samples must preserve the local repulsion direction at $v_{\rm a}$ while keeping their prediction-space gradient independent of $v$.

We retain the positive quadratic and linearize the negative branch at $v_{\rm a}$:
\begin{equation}
 \widetilde\ell_a(v)=
 \begin{cases}
   \ell_a(v),&a>0,\\
   \ell_a(v_{\rm a})+a\langle v_{\rm a}-u,v-v_{\rm a}\rangle,&a<0,\\
   0,&a=0.
 \end{cases}
 \label{eq:branch-objective}
\end{equation}
For a fixed sample, the resulting prediction-space update field is
\begin{equation}
 F_a(v)=-\nabla_v\widetilde\ell_a(v)=
 \begin{cases}
   a(u-v),&a\geq0,\\
   a(u-v_{\rm a}),&a<0.
 \end{cases}
 \label{eq:field}
\end{equation}
As illustrated in Figure~\ref{fig:method-roles}, positive updates track the current residual $u-v$ and vanish at $u$, whereas negative updates follow the fixed residual at $v_{\rm a}$. This branchwise objective is exactly Eq.~\eqref{eq:anchored} with a parameter-free gate:
\begin{equation}
 \widetilde\ell_a=\ell_{a,\omega(a)},\qquad
 \omega(a)=\max(-a,0).
 \label{eq:gate}
\end{equation}
The following proposition shows that the two requirements uniquely determine this gate within the quadratic-anchor family.

\begin{proposition}[Requirements characterize the gate]
\label{prop:aga}
Fix $a\ne0$ and a detached scalar coefficient $\omega$ in Eq.~\eqref{eq:anchored}.
For $a>0$, $u$ is the unique minimizer for every anchor $v_{\rm a}$ if and only if $\omega=0$.
For $a<0$, the gradient is independent of the current prediction $v$ if and only if $\omega=-a$. In the latter case, the objective forms the tangent upper bound
\begin{equation}
 \ell_{a,-a}(v)=\ell_a(v_{\rm a})+
 a\langle v_{\rm a}-u,v-v_{\rm a}\rangle\geq\ell_a(v),
 \label{eq:tangent}
\end{equation}
with constant update field $a(u-v_{\rm a})$.
\end{proposition}
\begin{proof}
The gradient is $(a+\omega)v-au-\omega v_{\rm a}$. At $v=u$, stationarity for every $v_{\rm a}$ requires $\omega(u-v_{\rm a})=0$, hence $\omega=0$, giving a unique minimum when $a>0$. Independence from $v$ requires $a+\omega=0$. Expanding $\ell_{a,-a}(v)$ gives Eq.~\eqref{eq:tangent}, whose gap from $\ell_a(v)$ is $(-a)K\geq0$.
\end{proof}

This tangent upper bound connects the negative branch to majorization-minimization (MM) and the concave-convex procedure (CCCP) \cite{mairal2015incremental,yuille2003concave}: at each step, $\ell_{a,-a}(v)$ majorizes the concave loss $\ell_a(v)$ with matching value and gradient at the expansion point $v_{\rm a}$, preventing repulsive gradients from growing as $v$ moves away from $u$.

By contrast, on the positive branch ($a>0$), $\ell_a(v)$ is already convex with its minimum at $u$. Keeping $\omega(a)=0$ retains the restoring curvature $aI\succ0$ that dampens updates as $v\to u$. Applying tangent linearization to positive samples as well instead sets $\omega=-a$ for all $a$. Whenever $v$ diverges from the lagging anchor $v_{\rm a}$ across updates, the positive update $a(u-v_{\rm a})$ no longer vanishes at $v=u$, differing from AGA by $F_{\rm AGA}-F_{\rm tangent}=a(v_{\rm a}-v)$ for $a>0$ and continuing to push $v$ past the target. Sections~\ref{sec:ablations} and~\ref{sec:ema-compatibility} examine the resulting stability and copy-update effects.

Because the gated anchor $\omega(a)=\max(-a,0)$ supplies a local negative expansion point rather than a retention penalty, it does not constrain global drift. We therefore complement reward matching with a separate constraint anchored at the frozen pretrained model $\theta_0$.


\subsection{Guided Reference Retention}
\label{sec:prior}

Although forward-process RL is often run CFG-free ($\gamma=1$) for efficiency~\cite{xue2025advantage,zheng2026diffusionnft}, classifier-free guidance (CFG) shapes the effective generation distribution of $\theta_0$ even when its weights are frozen~\cite{dhariwal2021diffusion,ho2022classifierfree}. Guidance therefore governs both the rollout trajectories that receive reward feedback and the pretrained behavior preserved by the reference.

\paragraph{Guidance selects a behavior distribution.}
At scale $\gamma$, CFG extrapolates conditional and unconditional predictions~\cite{ho2022classifierfree}:
\begin{equation}
 G_\theta^\gamma
 =\gamma v_\theta^c+(1-\gamma)v_\theta^\emptyset.
 \label{eq:guided-field}
\end{equation}
Paired with a solver and noise law, $G_\theta^\gamma$ defines the endpoint sampling policy $\pi_\theta^\gamma$. Let $\mu_\gamma=\pi_{\theta_0}^\gamma$ be the frozen pretrained endpoint distribution at scale $\gamma$, and let $\mu_1=\pi_{\theta_0}^1$ be the CFG-free baseline. We denote by $\gamma_\star$ the \emph{preferred guidance} representing the pretrained generation behavior we aim to retain, with $\mu_{\gamma_\star}$ as the frozen reference distribution.

\paragraph{Interpreting the reference choice.}
Consider an endpoint-level KL objective~\cite{levine2018inference,luo2024diffinstructpp,jin2025rlg}. For a fixed prompt, assume both reference densities are positive on the support of a candidate policy $q$ with finite log-ratio expectations. Defining the implicit CFG reward~\cite{luo2024diffinstructpp}
\begin{equation}
 R_{\rm cfg}^{\gamma}(x,c)
 =\tau\log\frac{\mu_\gamma(x\mid c)}{\mu_1(x\mid c)},
 \qquad \tau>0,
 \label{eq:implicit-guidance}
\end{equation}
rewrites the KL-regularized objective anchored at $\mu_\gamma$ in terms of the CFG-free baseline $\mu_1$:
\begin{equation}
\begin{aligned}
 \mathcal J_\gamma(q)
 &:=\mathbb E_q[r]-\tau D_{\rm KL}(q\|\mu_\gamma)\\
 &=\mathbb E_q[r+R_{\rm cfg}^{\gamma}]
   -\tau D_{\rm KL}(q\|\mu_1).
\end{aligned}
 \label{eq:guidance-reference-rewrite}
\end{equation}
Anchoring at $\mu_{\gamma_\star}$ thus augments the task reward $r$ with $R_{\rm cfg}^{\gamma_\star}$, favoring generations with higher likelihood under the guided pretrained prior, which is especially valuable when task rewards do not directly measure visual fidelity.

\paragraph{Guided-field retention.}
In forward-process RL, behavior retention is realized through velocity-space matching along noised trajectories~\cite{xue2025advantage,zhu2026crd,choi2026rvm}. Following CRD and RVM~\cite{zhu2026crd,choi2026rvm}, we match the composed guided field and use $\gamma_\star$ consistently across rollout, reference matching, and evaluation. Rollouts are sampled from the guided moving policy,
\begin{equation}
 x_0^i\sim\pi_{\theta_{\rm a}}^{\gamma_\star}(\cdot\mid c),
 \qquad r_i=r(x_0^i,c).
 \label{eq:rollout}
\end{equation}
Because $r(x_0^i,c)$ evaluates prompt-conditional alignment, the reward term updates the conditional velocity $v_\theta^c$ on re-noised states, while retention matches the current and frozen guided fields:
\begin{equation}
 \mathcal R_\star(\theta)=\frac12\mathbb E
 \big\|G_\theta^{\gamma_\star}(x_t,t,c)
       -G_{\theta_0}^{\gamma_\star}(x_t,t,c)\big\|^2.
 \label{eq:frozen}
\end{equation}
Writing $\delta_c=v_\theta^c-v_{\theta_0}^c$ and $\delta_\emptyset=v_\theta^\emptyset-v_{\theta_0}^\emptyset$, this penalty acts on the composed residual
\begin{equation}
 G_\theta^{\gamma_\star}-G_{\theta_0}^{\gamma_\star}
 =\gamma_\star\delta_c+(1-\gamma_\star)\delta_\emptyset.
 \label{eq:guided-residual}
\end{equation}
Unlike CFG-free regularization ($\gamma=1$), which leaves the extrapolated guided field unconstrained when $\gamma_\star>1$, or split-branch penalties that constrain $\delta_c$ and $\delta_\emptyset$ independently, $\mathcal R_\star(\theta)$ regularizes the exact linear combination executed during guided sampling while allowing the conditional and unconditional branches to co-adapt.

\paragraph{Choosing the preferred guidance.}
We measure whether pretrained generations across guidance scales align with human preference using Human Preference Score v3 (HPSv3)~\cite{ma2025hpsv3}. Scales with strong pretrained HPSv3 provide a better prior $\mu_{\gamma_\star}$, provided they avoid over-extrapolation over prolonged RL (Section~\ref{sec:guidance-scale}), motivating SD3.5-Medium's standard scale $\gamma_\star=4.5$~\cite{stabilityai2024sd35}.

\begin{algorithm}[t]
\caption{\textbf{Overview of AGA training.}}
\label{alg:aga-plus}
\small
\begin{algorithmic}[1]
\Require Pretrained $\theta_0$, reward $r$, $\gamma_\star$, EMA schedule $\{\rho_k\}$, $W,\lambda$.
\State Initialize $\theta\gets\theta_0$, $\theta_{\rm a}\gets\theta_0$, $k\gets0$, and freeze $\theta_0$.
\For{each rollout batch}
  \State Sample grouped $x_0^i\sim\pi_{\theta_{\rm a}}^{\gamma_\star}$ and compute $r_i,a_i$.
  \State Sample states via Eq.~\eqref{eq:path} and cache $v_{{\rm a},i},g_{0,i}$.
  \For{each optimizer step on minibatch $B$}
    \State Compute $v_i$ and detached $y_i$ using Eq.~\eqref{eq:target}.
    \State Update $\theta$ on $\widehat{\mathcal L}_{\rm match}+\lambda\widehat{\mathcal R}_\star$.
    \State Set $k\gets k+1$ and update $\theta_{\rm a}\gets\rho_k\theta_{\rm a}+(1-\rho_k)\theta$.
  \EndFor
\EndFor
\State \Return Guided policy $\pi_{\theta_{\rm a}}^{\gamma_\star}$.
\end{algorithmic}
\end{algorithm}

\subsection{Overview of the Training Procedure}
\label{sec:overview}
Combining local feedback with guided reference retention gives the complete training objective
\begin{equation}
 \mathcal L_{\rm AGA}(\theta)
 =\mathbb E_{x_0\sim\pi_{\theta_{\rm a}}^{\gamma_\star},\,t,\epsilon}
  \big[W(t)\widetilde\ell_a(v_\theta^c)\big]
  +\lambda\mathcal R_\star(\theta),
 \label{eq:complete}
\end{equation}
where $W(t)\geq0$ and $\lambda\geq0$. Here $\widetilde\ell_a$ preserves positive targets and linearizes negative repulsion, while $\mathcal R_\star$ anchors the composed guided field to $\theta_0$.

To implement the gated update with a standard regression loss, we construct the detached target
\begin{equation}
 y=\operatorname{sg}[v+F_a(v)],
 \label{eq:target}
\end{equation}
which interpolates toward the observed target via $v+a(u-v)$ for $a\geq0$ and steps away along the cached residual via $v+a(u-v_{\rm a})$ for $a<0$. Evaluating the standard squared loss $\tfrac12W(t)\|v-y\|^2$ (denoted $\widehat{\mathcal L}_{\rm match}$ on a batch) yields the exact first-order parameter gradient of $W(t)\widetilde\ell_a(v)$ without custom backward rules.

Before optimization step $k$, we cache the moving predictions $v_{{\rm a},i}$ and frozen guided predictions $g_{0,i}=\operatorname{sg}[G_{\theta_0}^{\gamma_\star}(x_{t_i},t_i,c_i)]$, and subsequently update $\theta_{\rm a}$ with EMA decay $\rho_k$. Algorithm~\ref{alg:aga-plus} summarizes the procedure.

\section{Experiments}
\label{sec:experiments}

\begin{table*}[!t]
 \centering
 \caption{\textbf{Task adaptation and image quality.} Full AGA achieves the highest task scores across all three benchmarks. The CFG-free variant (\emph{w/o Guided Policy}, $\gamma=1$) examines feedback design without guided adaptation. $\dagger$ denotes a checkpoint selected before pronounced image-quality degradation, and shaded cells mark the optimized task reward.}
 \label{tab:systems}
 \small
 \begin{tabular*}{\textwidth}{@{\extracolsep{\fill}}lcccccccc@{}}
\toprule
 & & \multicolumn{3}{c}{Task reward} & \multicolumn{4}{c}{DrawBench} \\
\cmidrule(lr){3-5}\cmidrule(lr){6-9}
Method & Step & GenEval & OCR & PickScore & ImgRwd & Aesthetic & CLIP & PickScore \\
\midrule
SD3.5M w/o CFG &  & 0.28 & 0.13 & 19.94 & -0.50 & 5.16 & 0.2379 & 20.74 \\
SD3.5M w/ CFG &  & 0.63 & 0.57 & 21.78 & 0.92 & 5.44 & 0.2868 & 22.45 \\
\midrule
\multicolumn{9}{c}{\textbf{Compositional Image Generation}} \\
\midrule
AWM & 200 & \rewardcell{\underline{0.94}} &  &  & \underline{0.75} & 5.30 & 0.2791 & \textbf{22.10} \\
DiffusionNFT & 200 & \rewardcell{0.82} &  &  & 0.48 & 5.19 & 0.2749 & 21.78 \\
RAM & 200 & \rewardcell{0.93} &  &  & 0.57 & \textbf{5.34} & \underline{0.2795} & 21.81 \\
AGA (Ours) & 200 & \rewardcell{\textbf{0.96}} &  &  & \textbf{0.97} & \underline{5.32} & \textbf{0.2864} & \underline{22.07} \\
\color{ablatedgray}\hspace{0.8em}w/o Guided Policy & \color{ablatedgray}200 & \color{ablatedgray}\rewardcell{\underline{0.94}} &  &  & \color{ablatedgray}0.54 & \color{ablatedgray}5.15 & \color{ablatedgray}0.2752 & \color{ablatedgray}21.91 \\
\midrule
\multicolumn{9}{c}{\textbf{Visual Text Rendering}} \\
\midrule
AWM & 300 &  & \rewardcell{0.79} &  & 0.80 & 5.27 & 0.2875 & 21.93 \\
DiffusionNFT$^{\dagger}$ & 200 &  & \rewardcell{0.89} &  & \textbf{0.89} & \underline{5.33} & \textbf{0.2890} & \underline{22.28} \\
RAM$^{\dagger}$ & 150 &  & \rewardcell{0.93} &  & 0.74 & \textbf{5.50} & 0.2832 & \textbf{22.30} \\
AGA (Ours) & 200 &  & \rewardcell{\textbf{0.97}} &  & 0.79 & 5.24 & \underline{0.2885} & 22.16 \\
\color{ablatedgray}\hspace{0.8em}w/o Guided Policy & \color{ablatedgray}300 &  & \color{ablatedgray}\rewardcell{\underline{0.94}} &  & \color{ablatedgray}\underline{0.83} & \color{ablatedgray}5.31 & \color{ablatedgray}0.2861 & \color{ablatedgray}22.23 \\
\midrule
\multicolumn{9}{c}{\textbf{Human Preference Alignment}} \\
\midrule
AWM & 400 &  &  & \rewardcell{22.69} & 1.02 & 5.97 & 0.2638 & 23.07 \\
DiffusionNFT & 400 &  &  & \rewardcell{22.38} & 0.96 & 5.89 & 0.2659 & 22.78 \\
RAM$^{\dagger}$ & 120 &  &  & \rewardcell{22.85} & 1.24 & 5.74 & \textbf{0.2854} & 23.21 \\
AGA (Ours) & 400 &  &  & \rewardcell{\textbf{23.57}} & \textbf{1.38} & \underline{6.16} & \underline{0.2790} & \underline{23.87} \\
\color{ablatedgray}\hspace{0.8em}w/o Guided Policy & \color{ablatedgray}400 &  &  & \color{ablatedgray}\rewardcell{\underline{23.52}} & \color{ablatedgray}\underline{1.29} & \color{ablatedgray}\textbf{6.30} & \color{ablatedgray}0.2707 & \color{ablatedgray}\textbf{23.89} \\
\bottomrule
\end{tabular*}

\end{table*}

\subsection{Experimental Setup}
\label{sec:settings}

We evaluate compositional generation on GenEval~\cite{ghosh2023geneval} (testing object counting, spatial relations, and attribute binding), visual text rendering on the OCR benchmark from Flow-GRPO~\cite{liu2025flowgrpo} (measuring character-level rendering accuracy), and human preference alignment on PickScore~\cite{kirstain2023pick}. Because optimizing a task reward can degrade visual realism or prompt fidelity outside the training distribution~\cite{kim2024confidence,li2026diffusionopd}, we also evaluate all adapted models on held-out DrawBench prompts~\cite{saharia2022photorealistic} using four complementary metrics: ImageReward~\cite{xu2023imagereward} and PickScore~\cite{kirstain2023pick} for human preference, Aesthetic Score~\cite{schuhmann2022aestheticpredictor} for visual appeal, and CLIP similarity~\cite{radford2021learning} for text-image alignment.

\paragraph{Training protocol.}
All methods fine-tune SD3.5-Medium~\cite{esser2024scaling,stabilityai2024sd35} at $512\times512$ resolution using rank-$32$ LoRA~\cite{hu2022lora} and AdamW with learning rate $3\times10^{-4}$. Each rollout batch samples $48$ prompts with $24$ images per prompt ($1{,}152$ images total) using a $10$-step DPM2 solver, subsamples $9$ re-noised timesteps per image, and performs one optimizer update; evaluation uses a $40$-step Euler solver. AGA uses quadratic timestep weighting $W(t)\propto t^2$ to emphasize higher-noise states where global layout forms, while AWM, DiffusionNFT, and RAM follow their respective official loss formulations (with $\beta_{\rm NFT}=0.1$ on OCR and $1.0$ on GenEval/PickScore for DiffusionNFT; details in Appendix~B). All methods share a frozen-reference penalty weight of $10^{-4}$ in coordinate-MSE units. Following prior work on reward over-optimization~\cite{kim2024confidence,li2026diffusionopd}, we report the final checkpoint when training remains stable and mark with $\dagger$ baseline checkpoints selected prior to severe quality degradation.

\paragraph{EMA schedules.}
All methods in Table~\ref{tab:systems} use shared scheduled EMA on GenEval and PickScore with decay $\rho_k=\min(0.01k,0.9)$, and shared fixed EMA with $\rho_k=0.9$ on OCR, where $k$ indexes optimizer updates. The moving copy $\theta_{\rm a}$ serves rollout, anchoring, and evaluation and is updated after each step $k$, while $\theta_0$ remains frozen.

\paragraph{Guidance scales.}
Following the standard protocols of prior work~\cite{xue2025advantage,zheng2026diffusionnft,bergmeister2026ram}, AWM, DiffusionNFT, RAM, and \emph{AGA w/o Guided Policy} use CFG-free ($\gamma=1$) rollout and reference matching, evaluating without CFG ($\gamma=1$) on GenEval and PickScore and at $\gamma=4.5$ on OCR. Full AGA applies SD3.5-Medium's standard scale $\gamma_\star=4.5$~\cite{stabilityai2024sd35} across rollout, reference matching, and evaluation on all three tasks.

\subsection{Improving the Reward--Quality Balance}
\label{sec:main-results}

Table~\ref{tab:systems} compares AGA against AWM~\cite{xue2025advantage}, DiffusionNFT~\cite{zheng2026diffusionnft}, and RAM~\cite{bergmeister2026ram}. We first evaluate the controlled CFG-free variant (\emph{w/o Guided Policy}, $\gamma=1$) under the exact training protocol of the baselines, then examine the complete guided recipe.

\begin{figure}[!t]
 \centering
 \includegraphics[width=\columnwidth]{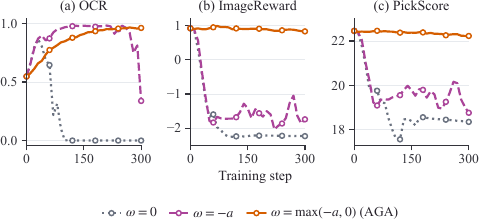}
 \caption{\textbf{Positive fitting preserves quality during reward learning.} With the same negative tangent branch, AGA sustains OCR accuracy and DrawBench quality, whereas the all-sign tangent rule ($\omega=-a$) loses quality before OCR declines.}
 \label{fig:anchor}
\end{figure}

\begin{table}[!t]
 \centering
  \vspace{-5mm}

 \caption{\textbf{Effects of anchor design.} Shaded cells mark the optimized reward (OCR), and dashes (--) denote training collapse.}
 \label{tab:shared-anchor}
 \small
 \setlength{\tabcolsep}{4pt}
 \renewcommand{\arraystretch}{0.95}
 \begin{tabular*}{\linewidth}{@{\extracolsep{\fill}}lccc@{}}
\toprule
\multirow{2}{*}{Anchor rule $\omega(a)$} & \multirow{2}{*}{OCR} & \multicolumn{2}{c}{DrawBench} \\
\cmidrule(lr){3-4}
 & & PickScore & ImgRwd \\
\midrule
\rowcolor{offsetgray}[0pt][0pt]
\multicolumn{4}{@{}l@{}}{\scriptsize\emph{Feedback components}} \\
$0$ & \rewardcell{--} & -- & -- \\
$-a$ & \rewardcell{0.34} & 18.76 & -1.74 \\
Positive-only & \rewardcell{--} & -- & -- \\
\addlinespace[0.5pt]
\rowcolor{offsetgray}[0pt][0pt]
\multicolumn{4}{@{}l@{}}{\scriptsize\emph{Affine rules}} \\
$0.1-a$ & \rewardcell{0.94} & 22.05 & 0.67 \\
$0.3-a$ & \rewardcell{0.87} & 21.62 & 0.39 \\
$1-a$ & \rewardcell{0.39} & 21.87 & 0.54 \\
$1$ & \rewardcell{0.61} & 21.18 & -0.01 \\
\addlinespace[0.5pt]
\rowcolor{offsetgray}[0pt][0pt]
\multicolumn{4}{@{}l@{}}{\scriptsize\emph{Offsets added to the gate}} \\
$0.1+\max(-a,0)$ & \rewardcell{0.90} & 22.08 & 0.74 \\
$0.5+\max(-a,0)$ & \rewardcell{0.73} & 21.11 & -0.04 \\
\addlinespace[0.5pt]
\rowcolor{offsetgray}[0pt][0pt]
\multicolumn{4}{@{}l@{}}{\scriptsize\emph{Positive targets with fixed negative feedback}} \\
$|a|$ & \rewardcell{0.91} & 22.06 & 0.74 \\
$\max(-a,0)$ (\textbf{AGA}) & \rewardcell{\textbf{0.94}} & \textbf{22.23} & \textbf{0.83} \\
\bottomrule
\end{tabular*}

 \vspace{-5mm}
\end{table}

Under matched CFG-free training ($\gamma=1$), \emph{AGA w/o Guided Policy} surpasses every baseline in task reward on OCR ($0.94$ vs.\ $0.93$) and PickScore ($23.52$ vs.\ $22.85$) while matching the top baseline on GenEval ($0.94$). On preference alignment, its gains extend to held-out DrawBench prompts, achieving the highest Aesthetic Score ($6.30$) and DrawBench PickScore ($23.89$) across all methods, alongside higher DrawBench ImageReward ($1.29$) than AWM ($1.02$), DiffusionNFT ($0.96$), and RAM ($1.24$). Moreover, whereas DiffusionNFT and RAM exhibit reward over-optimization on OCR or PickScore and require early stopping  before quality collapses, as also noted in DiffusionOPD~\cite{li2026diffusionopd}, AGA trains stably over the full horizon on every task.

Incorporating guided reference retention ($\gamma_\star=4.5$) further improves both task reward and visual quality: full AGA achieves the highest task scores on all three benchmarks ($0.96$ GenEval, $0.97$ OCR, and $23.57$ PickScore). The pretrained rows of Table~\ref{tab:systems} explain why guided retention is especially impactful on GenEval: CFG-free SD3.5-Medium ($\gamma=1$) starts with a weak visual prior (DrawBench ImageReward $-0.50$ vs.\ $0.92$ at $\gamma=4.5$). Because discrete object-composition rewards do not penalize visual artifacts, anchoring to the CFG-free prior yields lower ImageReward ($0.54$), whereas anchoring to $\mu_{\gamma_\star}$ retains the high-fidelity guided prior. Consequently, full AGA is the only method on GenEval to achieve large compositional gains ($0.63\to 0.96$) without losing pretrained DrawBench ImageReward ($0.97$ vs.\ $0.92$) or CLIP similarity ($0.2864$ vs.\ $0.2868$), whereas all CFG-free baselines drop ImageReward by $0.17$ to $0.44$. On OCR, full AGA reaches $0.97$ accuracy at step $200$ while preserving strong DrawBench CLIP similarity ($0.2885$) and PickScore ($22.16$). On preference alignment, full AGA similarly lifts ImageReward from $1.29$ to $1.38$ and CLIP similarity from $0.2707$ to $0.2790$ over its CFG-free variant while maintaining strong DrawBench PickScore ($23.87$).

\subsection{Effects of Asymmetric Feedback}
\label{sec:ablations}

We next examine which elements of the feedback rule sustain adaptation. Figure~\ref{fig:anchor} and Table~\ref{tab:shared-anchor} compare anchor rules $\omega(a)$ on OCR under CFG-free training ($\gamma=1$), evaluation guidance $4.5$, and identical frozen-reference regularization.

Leaving negative samples unanchored ($\omega=0$) or omitting negative feedback (\emph{Positive-only}) both collapse training (Table~\ref{tab:shared-anchor}; $\omega=0$ in Figure~\ref{fig:anchor}), confirming both halves of Sections~\ref{sec:policy}--\ref{sec:aga}. Under $\omega=0$, concave negative curvature makes repulsion grow with $\|v-u\|$, causing self-amplifying divergence. Conversely, dropping negative feedback eliminates repulsion against poor rollouts, so positive-only self-fitting reinforces failing rollout modes and collapses. Stable adaptation thus requires keeping negative repulsion while removing its curvature via tangent linearization at $v_{\rm a}$.

\paragraph{Positive target preservation prevents quality collapse.}
Holding AGA's negative tangent branch fixed, we first replace positive fitting with the all-sign tangent rule $\omega=-a$ (RAM's moving-anchor component). Positive updates then follow the cached direction $a(u-v_{\rm a})$ rather than the live residual $a(u-v)$ and need not vanish at $u$. Figure~\ref{fig:anchor} reveals a striking two-phase failure mode under $\omega=-a$: during the first $100$ steps, OCR accuracy rises rapidly alongside AGA (Figure~\ref{fig:anchor}a), while held-out DrawBench ImageReward and PickScore plummet from the onset of training (Figures~\ref{fig:anchor}b,c); once visual quality has degraded, OCR accuracy itself peaks and collapses to $0.34$ (with ImageReward $-1.74$, vs.\ $0.94$ and $0.83$ for AGA). High early task reward can thus mask severe representation degradation when positive updates are not self-limiting at $u$.

We also test whether a contracted unique positive target suffices. Setting $\omega=|a|$ keeps AGA's negative tangent branch unchanged and adds a positive anchor of weight $a$, shifting the positive fitting point to $(u+v_{\rm a})/2$. At the same endpoint, this contraction lowers OCR to $0.91$ and ImageReward to $0.74$ (Table~\ref{tab:shared-anchor}), showing that pulling positive targets toward the moving copy degrades adaptation even when a unique minimum exists.

\paragraph{Branch decoupling outperforms tuned affine rules.}
We next test whether coefficient tuning can match AGA's decoupled gate. For $\omega=\beta-a$, varying $\beta$ jointly alters negative curvature $\beta$ and positive target scaling $s(b)=b/\beta$. At $\beta=1$ (FlowAWR's rule $1-a$), contracting positive steps for $b<1$ limits OCR to $0.39$ (ImageReward $0.54$). Lowering $\beta$ to $0.3$ and $0.1$ improves OCR to $0.87$ and $0.94$, but positive target rescaling at $\beta=0.1$ still trails AGA on both DrawBench PickScore ($22.05$ vs.\ $22.23$) and ImageReward ($0.67$ vs.\ $0.83$), while a constant anchor ($\omega=1$, with $s(b)=b/(b+1)$) reduces OCR to $0.61$ and ImageReward to $-0.01$. Adding a positive offset $\omega=\delta+\max(-a,0)$ similarly restores positive curvature on negative samples at the cost of contracting positive targets to $s(b)=b/(b+\delta)<1$, lowering OCR/ImageReward to $0.90/0.74$ at $\delta=0.1$ and $0.73/{-}0.04$ at $\delta=0.5$. Neither tuning affine coefficients nor adding a gate offset matches AGA's parameter-free gate.

\subsection{Gains Persist across EMA Protocols}
\label{sec:ema-compatibility}

\begin{figure}[!t]
 \centering
 \includegraphics[width=\columnwidth]{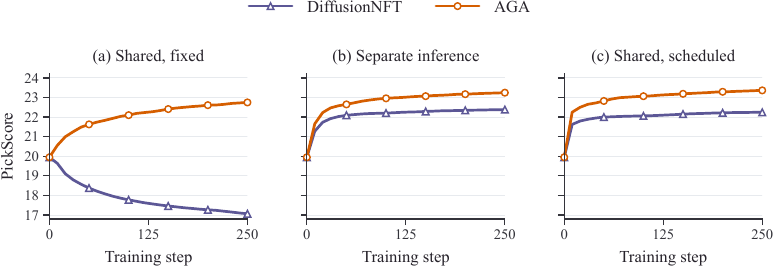}
  \vspace{-3mm}

 \caption{\textbf{Sensitivity to policy-copy updates on PickScore.} AGA improves across all three protocols. Updating the policy more rapidly early in training restores DiffusionNFT's learning, but AGA retains higher reward with either evaluation-copy choice.}
 \label{fig:ema-compatibility}
\end{figure}

Because affine rules couple positive targets to $v_{\rm a}$, their updates depend on how closely $\theta_{\rm a}$ tracks $\theta$. We test sensitivity to this copy-update schedule by comparing AGA with adaptive DiffusionNFT ($\beta=1$) on PickScore under matched CFG-free training ($\gamma=1$) across three EMA protocols adapted from RAM~\cite{bergmeister2026ram} (Figure~\ref{fig:ema-compatibility}): (a) one copy with fixed decay $0.9$ for rollout, anchoring, and evaluation; (b) RAM's two-copy setup, with scheduled decay $\min(0.01k,0.9)$ for rollout/anchoring and fixed decay $0.9$ for evaluation; and (c) one copy with scheduled decay $\min(0.01k,0.9)$ for all three roles (used in Table~\ref{tab:systems}).

DiffusionNFT deteriorates under the slower fixed-EMA anchor in (a), dropping to $17.06$ PickScore, and relies on the fast early anchor updates in (b)--(c) to recover learning ($22.39$ and $22.25$). By contrast, because AGA's positive updates $a(u-v)$ are independent of $v_{\rm a}$, AGA improves steadily across all three protocols, reaching $22.75$ in (a), $23.25$ in (b), and $23.37$ in (c), and retains higher PickScore under either evaluation-copy choice.

\subsection{Guidance and Quality Retention}
\label{sec:reward-recipes}
\label{sec:guidance-scale}

Finally, we examine guidance as part of the full AGA recipe, comparing guided OCR-only training with explicit reward mixing and assessing guidance selection over adaptation.

\begin{table}[!t]
 \centering
 \caption{\textbf{Guided adaptation versus explicit reward mixing.} OCR-only guided AGA improves both text accuracy and DrawBench preference scores over CFG-free training.}
 \label{tab:explicit-reward}
 \small
 \setlength{\tabcolsep}{4pt}
 \renewcommand{\arraystretch}{0.95}
 \begin{tabular*}{\linewidth}{@{\extracolsep{\fill}}lcccc@{}}
\toprule
\multirow{2}{*}{Explicit reward} & \multirow{2}{*}{CFG} & \multirow{2}{*}{OCR} & \multicolumn{2}{c}{DrawBench} \\
\cmidrule(lr){4-5}
 & & & PickScore & ImgRwd \\
\midrule
OCR + PickScore & $1$ & \rewardcell{0.94} & \rewardcell{21.88} & 0.58 \\
OCR & $4.5$ & \rewardcell{0.97} & 22.16 & 0.79 \\
\bottomrule
\end{tabular*}

  \vspace{-3mm}

\end{table}

Section~\ref{sec:prior} shows that anchoring to $\mu_{\gamma_\star}$ supplies an implicit preference reward $R_{\rm cfg}^{\gamma_\star}$. Table~\ref{tab:explicit-reward} tests whether guided OCR-only AGA ($\gamma_\star=4.5$) can retain preference quality compared with CFG-free AGA ($\gamma=1$) trained on an explicit mixture of OCR reward and PickScore (scaled by $1/26$ to place both reward terms on a comparable scale), both evaluated at CFG $4.5$. Mixing two rewards into $a$ forces one scalar advantage to trade off text legibility against general preference under the CFG-free prior $\mu_1$, whereas guided OCR-only training keeps $a$ dedicated to text accuracy while regularizing toward $\mu_{\gamma_\star}$, improving OCR ($0.94$ to $0.97$), DrawBench PickScore ($21.88$ to $22.16$), and ImageReward ($0.58$ to $0.79$) without an external preference reward.

\begin{figure}[!t]
 \centering
 \includegraphics[width=\columnwidth]{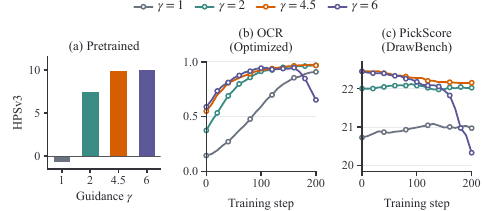}
  \vspace{-5mm}

 \caption{\textbf{Pretrained quality guides scale selection.} Guidance scales with high pretrained HPSv3 ($\gamma=4.5,6$) achieve stronger early OCR learning and DrawBench PickScore than $\gamma=1,2$, while $\gamma_\star=4.5$ avoids late drops from over-extrapolation at $\gamma=6$.}
 \label{fig:guidance-scale}
\end{figure}

\paragraph{Pretrained quality informs guidance selection.}
Sweeping rollout, reference, and evaluation guidance over $\{1,2,4.5,6\}$ on OCR (Figure~\ref{fig:guidance-scale}) supports choosing $\gamma_\star$ via pretrained Human Preference Score v3 (HPSv3)~\cite{ma2025hpsv3} (Section~\ref{sec:prior}): pretrained HPSv3 rises sharply from $-0.76$ ($\gamma=1$) to $7.46$ ($\gamma=2$) and peaks at $\gamma=4.5$ ($9.96$) and $6$ ($10.04$) (Figure~\ref{fig:guidance-scale}a). Matching this preference ranking, $\gamma=4.5$ and $6$ deliver stronger early OCR gains and DrawBench PickScore than $\gamma=1,2$ (Figures~\ref{fig:guidance-scale}b,c). However, $\gamma=6$ adds minimal HPSv3 over $4.5$ ($10.04$ vs.\ $9.96$) while over-extrapolating predictions, causing oversaturation and sharp drops after $150$ steps, whereas $\gamma_\star=4.5$ stays stable.

\section{Conclusion}
\label{sec:conclusion}

Moving-policy anchors introduced to stabilize negative feedback in forward-process RL can inadvertently alter what positive feedback teaches a diffusion model to fit. We formalize this conflict as \emph{anchor coupling}, proving that within the local quadratic family, affine moving-anchor rules cannot eliminate negative curvature while preserving every positive regression target. Advantage-Gated Adaptation (AGA) resolves this coupling by retaining self-limiting target regression on positive samples and linearizing the concave objective at the moving-policy prediction on negative samples. Within the quadratic-anchor family, these requirements uniquely determine the parameter-free gate $\omega(a)=\max(-a,0)$. Complementing local feedback with a frozen pretrained reference regularized at its operating guidance scale separates step-level stabilization from long-horizon behavior retention. Across compositional generation, text rendering, and preference alignment, our results show that displacing positive targets degrades visual fidelity and heightens sensitivity to policy-copy schedules, establishing positive target preservation alongside negative-curvature control as a core principle for diffusion RL.

\label{paper:body-end}
\FloatBarrier
{
\small
\bibliographystyle{ieeenat_fullname}
\bibliography{references}

\begin{thebibliography}{38}
\providecommand{\natexlab}[1]{#1}
\providecommand{\url}[1]{\texttt{#1}}
\expandafter\ifx\csname urlstyle\endcsname\relax
  \providecommand{\doi}[1]{doi: #1}\else
  \providecommand{\doi}{doi: \begingroup \urlstyle{rm}\Url}\fi

\bibitem[Bergmeister et~al.(2026)Bergmeister, Jegelka, N{\"u}sken,
  Domingo-Enrich, and Pidstrigach]{bergmeister2026ram}
Andreas Bergmeister, Stefanie Jegelka, Nikolas N{\"u}sken, Carles
  Domingo-Enrich, and Jakiw Pidstrigach.
\newblock Reinforce adjoint matching: Scaling {RL} post-training of diffusion
  and flow-matching models.
\newblock \emph{arXiv preprint arXiv:2605.10759}, 2026.

\bibitem[Black et~al.(2024)Black, Janner, Du, Kostrikov, and
  Levine]{black2024training}
Kevin Black, Michael Janner, Yilun Du, Ilya Kostrikov, and Sergey Levine.
\newblock Training diffusion models with reinforcement learning.
\newblock In \emph{International Conference on Learning Representations}, 2024.

\bibitem[Choi et~al.(2026)Choi, Guo, Zhu, Vahdat, Tao, Berner, and
  Chen]{choi2026rvm}
Jaemoo Choi, Wei Guo, Yuchen Zhu, Arash Vahdat, Molei Tao, Julius Berner, and
  Yongxin Chen.
\newblock Scaling reinforcement learning for diffusion models via velocity
  matching.
\newblock \emph{arXiv preprint arXiv:2608.23664v2}, 2026.

\bibitem[Clark et~al.(2024)Clark, Vicol, Swersky, and Fleet]{clark2024draft}
Kevin Clark, Paul Vicol, Kevin Swersky, and David Fleet.
\newblock Directly fine-tuning diffusion models on differentiable rewards.
\newblock In \emph{International Conference on Learning Representations}, 2024.

\bibitem[Dhariwal and Nichol(2021)]{dhariwal2021diffusion}
Prafulla Dhariwal and Alex Nichol.
\newblock Diffusion models beat {GAN}s on image synthesis.
\newblock In \emph{Advances in Neural Information Processing Systems}, 2021.

\bibitem[Esser et~al.(2024)Esser, Kulal, Blattmann, Entezari, M{\"u}ller,
  Saini, Levi, Lorenz, Sauer, Boesel, Podell, Dockhorn, English, and
  Rombach]{esser2024scaling}
Patrick Esser, Sumith Kulal, Andreas Blattmann, Rahim Entezari, Jonas
  M{\"u}ller, Harry Saini, Yam Levi, Dominik Lorenz, Axel Sauer, Frederic
  Boesel, Dustin Podell, Tim Dockhorn, Zion English, and Robin Rombach.
\newblock Scaling rectified flow transformers for high-resolution image
  synthesis.
\newblock In \emph{International Conference on Machine Learning}, pages
  12606--12633. PMLR, 2024.

\bibitem[Fan et~al.(2023)Fan, Watkins, Du, Liu, Ryu, Boutilier, Abbeel,
  Ghavamzadeh, Lee, and Lee]{fan2023dpok}
Ying Fan, Olivia Watkins, Yuqing Du, Hao Liu, Moonkyung Ryu, Craig Boutilier,
  Pieter Abbeel, Mohammad Ghavamzadeh, Kangwook Lee, and Kimin Lee.
\newblock {DPOK}: Reinforcement learning for fine-tuning text-to-image
  diffusion models.
\newblock In \emph{Advances in Neural Information Processing Systems}, pages
  79858--79885, 2023.

\bibitem[Fu et~al.(2026)Fu, He, Shang, Ma, Mu, Wang, Liu, and
  Fu]{fu2026flowawr}
Zheming Fu, Ruizhe He, Wei Shang, Xiaoxiao Ma, Lianrui Mu, Lei Wang, Chang Liu,
  and Siming Fu.
\newblock {FlowAWR}: Online adaptive flow reinforcement via advantage-weighted
  rectification.
\newblock \emph{arXiv preprint arXiv:2606.30376}, 2026.

\bibitem[Ghosh et~al.(2023)Ghosh, Hajishirzi, and Schmidt]{ghosh2023geneval}
Dhruba Ghosh, Hannaneh Hajishirzi, and Ludwig Schmidt.
\newblock {GenEval}: An object-focused framework for evaluating text-to-image
  alignment.
\newblock In \emph{Advances in Neural Information Processing Systems}, pages
  52132--52152, 2023.

\bibitem[Ho and Salimans(2022)]{ho2022classifierfree}
Jonathan Ho and Tim Salimans.
\newblock Classifier-free diffusion guidance.
\newblock \emph{arXiv preprint arXiv:2207.12598}, 2022.

\bibitem[Ho et~al.(2020)Ho, Jain, and Abbeel]{ho2020denoising}
Jonathan Ho, Ajay Jain, and Pieter Abbeel.
\newblock Denoising diffusion probabilistic models.
\newblock In \emph{Advances in Neural Information Processing Systems}, pages
  6840--6851, 2020.

\bibitem[Hu et~al.(2022)Hu, Shen, Wallis, Allen-Zhu, Li, Wang, Wang, and
  Chen]{hu2022lora}
Edward~J. Hu, Yelong Shen, Phillip Wallis, Zeyuan Allen-Zhu, Yuanzhi Li, Shean
  Wang, Lu Wang, and Weizhu Chen.
\newblock {LoRA}: Low-rank adaptation of large language models.
\newblock In \emph{International Conference on Learning Representations}, 2022.

\bibitem[Jin et~al.(2025)Jin, Qiu, Liu, Diao, Qiao, Ding, Lamb, and
  Qiu]{jin2025rlg}
Luozhijie Jin, Zijie Qiu, Jie Liu, Zijie Diao, Lifeng Qiao, Ning Ding, Alex
  Lamb, and Xipeng Qiu.
\newblock Inference-time alignment control for diffusion models with
  reinforcement learning guidance.
\newblock \emph{arXiv preprint arXiv:2508.21016}, 2025.

\bibitem[Kim et~al.(2024)Kim, Jeong, An, Ghavamzadeh, Dvijotham, Shin, and
  Lee]{kim2024confidence}
Kyuyoung Kim, Jongheon Jeong, Minyong An, Mohammad Ghavamzadeh, Krishnamurthy
  Dvijotham, Jinwoo Shin, and Kimin Lee.
\newblock Confidence-aware reward optimization for fine-tuning text-to-image
  models.
\newblock In \emph{International Conference on Learning Representations}, 2024.

\bibitem[Kirstain et~al.(2023)Kirstain, Polyak, Singer, Matiana, Penna, and
  Levy]{kirstain2023pick}
Yuval Kirstain, Adam Polyak, Uriel Singer, Shahbuland Matiana, Joe Penna, and
  Omer Levy.
\newblock {Pick-a-Pic}: An open dataset of user preferences for text-to-image
  generation.
\newblock In \emph{Advances in Neural Information Processing Systems}, pages
  36652--36663, 2023.

\bibitem[Kveton et~al.(2026)Kveton, Rao, Mukherjee, Singh, and
  Lai]{kveton2026advantageflow}
Branislav Kveton, Anup Rao, Subhojyoti Mukherjee, Krishna~Kumar Singh, and
  Viet~Dac Lai.
\newblock {AdvantageFlow}: Regularized advantage-weighted {RL} in flow models.
\newblock \emph{arXiv preprint arXiv:2605.26013v2}, 2026.

\bibitem[Levine(2018)]{levine2018inference}
Sergey Levine.
\newblock Reinforcement learning and control as probabilistic inference:
  Tutorial and review.
\newblock \emph{arXiv preprint arXiv:1805.00909}, 2018.

\bibitem[Li et~al.(2026)Li, Yu, Jiang, Wei, Xing, Li, Chu, Zhang, Liu, and
  Wu]{li2026diffusionopd}
Quanhao Li, Junqiu Yu, Kaixun Jiang, Yujie Wei, Zhen Xing, Pandeng Li, Ruihang
  Chu, Shiwei Zhang, Yu Liu, and Zuxuan Wu.
\newblock {DiffusionOPD}: A unified perspective of on-policy distillation in
  diffusion models.
\newblock \emph{arXiv preprint arXiv:2605.15055}, 2026.

\bibitem[Lipman et~al.(2023)Lipman, Chen, Ben-Hamu, Nickel, and
  Le]{lipman2023flow}
Yaron Lipman, Ricky T.~Q. Chen, Heli Ben-Hamu, Maximilian Nickel, and Matt Le.
\newblock Flow matching for generative modeling.
\newblock In \emph{International Conference on Learning Representations}, 2023.

\bibitem[Liu et~al.(2025{\natexlab{a}})Liu, Gao, Liu, Du, Li, Wu, Jin, Cao,
  Zhang, Li, and Hoi]{liu2025decoupleddmd}
Dongyang Liu, Peng Gao, David Liu, Ruoyi Du, Zhen Li, Qilong Wu, Xin Jin, Sihan
  Cao, Shifeng Zhang, Hongsheng Li, and Steven Hoi.
\newblock Decoupled {DMD}: {CFG} augmentation as the spear, distribution
  matching as the shield.
\newblock \emph{arXiv preprint arXiv:2511.22677}, 2025{\natexlab{a}}.

\bibitem[Liu et~al.(2025{\natexlab{b}})Liu, Liu, Liang, Li, Liu, Wang, Wan,
  Zhang, and Ouyang]{liu2025flowgrpo}
Jie Liu, Gongye Liu, Jiajun Liang, Yangguang Li, Jiaheng Liu, Xintao Wang,
  Pengfei Wan, Di Zhang, and Wanli Ouyang.
\newblock {Flow-GRPO}: Training flow matching models via online {RL}.
\newblock In \emph{Advances in Neural Information Processing Systems}, pages
  40783--40818, 2025{\natexlab{b}}.

\bibitem[Luo(2024)]{luo2024diffinstructpp}
Weijian Luo.
\newblock {Diff-Instruct++}: Training one-step text-to-image generator model to
  align with human preferences.
\newblock \emph{Transactions on Machine Learning Research}, 2024.

\bibitem[Ma et~al.(2025)Ma, Shui, Wu, Sun, and Li]{ma2025hpsv3}
Yuhang Ma, Yun-Hao Shui, Xiaoshi Wu, Keqiang Sun, and Hongsheng Li.
\newblock {HPSv3}: Towards wide-spectrum human preference score.
\newblock In \emph{Proceedings of the IEEE/CVF International Conference on
  Computer Vision}, pages 15086--15095, 2025.

\bibitem[Mairal(2015)]{mairal2015incremental}
Julien Mairal.
\newblock Incremental majorization-minimization optimization with application
  to large-scale machine learning.
\newblock \emph{SIAM Journal on Optimization}, 25\penalty0 (2):\penalty0
  829--855, 2015.

\bibitem[Radford et~al.(2021)Radford, Kim, Hallacy, Ramesh, Goh, Agarwal,
  Sastry, Askell, Mishkin, Clark, Krueger, and Sutskever]{radford2021learning}
Alec Radford, Jong~Wook Kim, Chris Hallacy, Aditya Ramesh, Gabriel Goh,
  Sandhini Agarwal, Girish Sastry, Amanda Askell, Pamela Mishkin, Jack Clark,
  Gretchen Krueger, and Ilya Sutskever.
\newblock Learning transferable visual models from natural language
  supervision.
\newblock In \emph{Proceedings of the 38th International Conference on Machine
  Learning}, pages 8748--8763. PMLR, 2021.

\bibitem[Rombach et~al.(2022)Rombach, Blattmann, Lorenz, Esser, and
  Ommer]{rombach2022high}
Robin Rombach, Andreas Blattmann, Dominik Lorenz, Patrick Esser, and Bj{\"o}rn
  Ommer.
\newblock High-resolution image synthesis with latent diffusion models.
\newblock In \emph{Proceedings of the IEEE/CVF Conference on Computer Vision
  and Pattern Recognition}, 2022.

\bibitem[Saharia et~al.(2022)Saharia, Chan, Saxena, Li, Whang, Denton,
  Ghasemipour, Ayan, Mahdavi, Gontijo-Lopes, Salimans, Ho, Fleet, and
  Norouzi]{saharia2022photorealistic}
Chitwan Saharia, William Chan, Saurabh Saxena, Lala Li, Jay Whang, Emily
  Denton, Seyed Kamyar~Seyed Ghasemipour, Burcu~Karagol Ayan, S.~Sara Mahdavi,
  Raphael Gontijo-Lopes, Tim Salimans, Jonathan Ho, David~J. Fleet, and
  Mohammad Norouzi.
\newblock Photorealistic text-to-image diffusion models with deep language
  understanding.
\newblock In \emph{Advances in Neural Information Processing Systems}, pages
  36479--36494, 2022.

\bibitem[Schuhmann(2022)]{schuhmann2022aestheticpredictor}
Christoph Schuhmann.
\newblock {CLIP+MLP} aesthetic score predictor.
\newblock
  \url{https://github.com/christophschuhmann/improved-aesthetic-predictor},
  2022.
\newblock GitHub repository.

\bibitem[Schulman et~al.(2017)Schulman, Wolski, Dhariwal, Radford, and
  Klimov]{schulman2017proximal}
John Schulman, Filip Wolski, Prafulla Dhariwal, Alec Radford, and Oleg Klimov.
\newblock Proximal policy optimization algorithms.
\newblock \emph{arXiv preprint arXiv:1707.06347}, 2017.

\bibitem[{Stability AI}(2024)]{stabilityai2024sd35}
{Stability AI}.
\newblock {Stable Diffusion 3.5 Medium}.
\newblock Model card, 2024.

\bibitem[Wallace et~al.(2024)Wallace, Dang, Rafailov, Zhou, Lou, Purushwalkam,
  Ermon, Xiong, Joty, and Naik]{wallace2024diffusiondpo}
Bram Wallace, Meihua Dang, Rafael Rafailov, Linqi Zhou, Aaron Lou, Senthil
  Purushwalkam, Stefano Ermon, Caiming Xiong, Shafiq Joty, and Nikhil Naik.
\newblock Diffusion model alignment using direct preference optimization.
\newblock In \emph{Proceedings of the IEEE/CVF Conference on Computer Vision
  and Pattern Recognition}, pages 8228--8238, 2024.

\bibitem[Xu et~al.(2023)Xu, Liu, Wu, Tong, Li, Ding, Tang, and
  Dong]{xu2023imagereward}
Jiazheng Xu, Xiao Liu, Yuchen Wu, Yuxuan Tong, Qinkai Li, Ming Ding, Jie Tang,
  and Yuxiao Dong.
\newblock {ImageReward}: Learning and evaluating human preferences for
  text-to-image generation.
\newblock In \emph{Advances in Neural Information Processing Systems}, pages
  15903--15935, 2023.

\bibitem[Xue et~al.(2025{\natexlab{a}})Xue, Ge, Zhang, Li, and
  Ma]{xue2025advantage}
Shuchen Xue, Chongjian Ge, Shilong Zhang, Yichen Li, and Zhi-Ming Ma.
\newblock Advantage weighted matching: Aligning {RL} with pretraining in
  diffusion models.
\newblock \emph{arXiv preprint arXiv:2509.25050}, 2025{\natexlab{a}}.

\bibitem[Xue et~al.(2025{\natexlab{b}})Xue, Wu, Gao, Kong, Zhu, Chen, Liu, Liu,
  Guo, Huang, and Luo]{xue2025dancegrpo}
Zeyue Xue, Jie Wu, Yu Gao, Fangyuan Kong, Lingting Zhu, Mengzhao Chen, Zhiheng
  Liu, Wei Liu, Qiushan Guo, Weilin Huang, and Ping Luo.
\newblock {DanceGRPO}: Unleashing {GRPO} on visual generation.
\newblock \emph{arXiv preprint arXiv:2505.07818}, 2025{\natexlab{b}}.

\bibitem[Yang et~al.(2024)Yang, Chen, and Zhou]{yang2024dense}
Shentao Yang, Tianqi Chen, and Mingyuan Zhou.
\newblock A dense reward view on aligning text-to-image diffusion with
  preference.
\newblock In \emph{International Conference on Machine Learning}, pages
  55998--56032. PMLR, 2024.

\bibitem[Yuille and Rangarajan(2003)]{yuille2003concave}
Alan~L. Yuille and Anand Rangarajan.
\newblock The concave-convex procedure.
\newblock \emph{Neural Computation}, 15\penalty0 (4):\penalty0 915--936, 2003.

\bibitem[Zheng et~al.(2026)Zheng, Chen, Ye, Wang, Zhang, Jiang, Su, Ermon, Zhu,
  and Liu]{zheng2026diffusionnft}
Kaiwen Zheng, Huayu Chen, Haotian Ye, Haoxiang Wang, Qinsheng Zhang, Kai Jiang,
  Hang Su, Stefano Ermon, Jun Zhu, and Ming-Yu Liu.
\newblock {DiffusionNFT}: Online diffusion reinforcement with forward process.
\newblock In \emph{International Conference on Learning Representations}, 2026.

\bibitem[Zhu et~al.(2026)Zhu, Wang, Lathuili{\`e}re, and
  Kalogeiton]{zhu2026crd}
Yuanzhi Zhu, Xi Wang, St{\'e}phane Lathuili{\`e}re, and Vicky Kalogeiton.
\newblock Diffusion reinforcement learning via centered reward distillation.
\newblock \emph{arXiv preprint arXiv:2603.14128}, 2026.

\end{thebibliography}
}
\clearpage
\section*{Supplementary Material}
\appendix
\setcounter{figure}{0}
\setcounter{table}{0}
\setcounter{equation}{0}
\renewcommand{\thefigure}{S\arabic{figure}}
\renewcommand{\thetable}{S\arabic{table}}
\renewcommand{\theequation}{S\arabic{equation}}
\renewcommand{\theHfigure}{supp.\arabic{figure}}
\renewcommand{\theHtable}{supp.\arabic{table}}
\renewcommand{\theHequation}{supp.\arabic{equation}}
\section{Supplementary Derivations}

\subsection{Proof of the gated objective}

For a fixed sample and detached coefficients $a,\omega$, the anchored objective
\begin{equation}
 \ell_{a,\omega}(v)=\frac a2\|v-u\|^2+
                    \frac\omega2\|v-v_{\rm a}\|^2
\end{equation}
has gradient $(a+\omega)v-au-\omega v_{\rm a}$. For $a>0$, stationarity at $u$ for every $v_{\rm a}$ requires $\omega(u-v_{\rm a})=0$, hence $\omega=0$. The Hessian is then $aI\succ0$. For $a<0$, independence of the gradient from $v$ requires $a+\omega=0$, hence $\omega=-a$. These conditions give $\omega(a)=\max(-a,0)$, with zero contribution at $a=0$.

For $a<0$, expanding the objective gives
\begin{align}
 \ell_{a,-a}(v)
 &=a\langle v,v_{\rm a}-u\rangle
   +\frac a2(\|u\|^2-\|v_{\rm a}\|^2)\\
 &=\ell_a(v_{\rm a})+
   a\langle v_{\rm a}-u,v-v_{\rm a}\rangle,
\end{align}
where $\ell_a(v)=\tfrac a2\|v-u\|^2$. The gap
\begin{equation}
 \ell_{a,-a}(v)-\ell_a(v)
       =\frac{-a}{2}\|v-v_{\rm a}\|^2\geq0
\end{equation}
shows that this affine objective is a tangent upper bound, with matching value and gradient at $v=v_{\rm a}$.

\subsection{Deriving the moving-anchor rules}
\label{sec:supp-anchor-rules}

We derive the local quadratic forms used in the main comparison. We hold the sampled state, target, reference predictions, and scalar coefficients fixed when differentiating. Write $D=\tfrac12\|v-u\|^2$, $K=\tfrac12\|v-v_{\rm a}\|^2$, $R=\tfrac12\|v-v_{\rm ref}\|^2$, and $\Delta=u-v_{\rm a}$. The coefficient $a$ denotes each method's effective signed matching weight. Positive factors shared by the two matching terms are absorbed into the sample weight. Frozen-reference terms are recorded separately from the moving-anchor component.

For $\ell=aD+\omega(a)K$ with $a+\omega(a)>0$, completing the square gives
\begin{equation}
 \ell=\frac{a+\omega(a)}2
 \left\|v-\frac{au+\omega(a)v_{\rm a}}{a+\omega(a)}\right\|^2+C,
\end{equation}
where $C$ is independent of $v$. The two entries in each main-table row follow by substitution:
\begin{equation}
 \begin{aligned}
 \kappa_-(b)&=\omega(-b)-b,\\
 s_+(b)&=\frac{b}{b+\omega(b)},\qquad b>0.
 \end{aligned}
 \label{eq:supp-anchor-row-check}
\end{equation}
The second expression describes a unique minimum when its denominator is positive. At zero curvature, the matching component is affine and its feedback field is $a\Delta$. The reductions retain the half-squared-norm convention and prediction-independent term $C$ defined above. Substitution into Eq.~\eqref{eq:supp-anchor-row-check} gives the properties in the main table.

\paragraph{Coefficient and reference conventions.}
AdvantageFlow's $\beta(c)=\max_i|a_i|+\gamma_0$, with $\gamma_0>0$, is shared by the samples of a prompt during each batch. It gives $\beta(c)-b\geq\gamma_0$ and $s_+(b)<1/2$. FlowAWR's theoretical centered weight in Eq.~(9) is $\mathcal A=\exp(R/\gamma)/\phi-1>-1$. Its practical rule-based-reward configurations additionally rescale this weight by the within-group standard deviation, so the effective $a$ can extend below $-1$ \cite[Appendix~C.4]{fu2026flowawr}. The quadratic identity applies to either effective weight.

The RAM paper's Eq.~(17) uses the current prediction inside the detached target, $v_{\rm ref}+a(u-v)$, giving a first-order field update-equivalent to $aD+R$ \cite{bergmeister2026ram}. The released implementation instead uses $v_{\rm a}$ in that residual, yielding $R+a(D-K)+C$. Its complete squared regression has curvature one and target $v_{\rm ref}+a\Delta$. After separating $R$, the moving-anchor component $a(D-K)$ matches AGA's negative branch but also makes the positive branch affine.

RVM allows a general coefficient function $\beta(a)$ and a configurable anchor prediction, encompassing both affine and sign-gated rules \cite{choi2026rvm}.

\subsection{Parameter gradients and detached targets}

We use Euclidean norms on $d$ velocity coordinates and write the coordinate-wise mean squared error as $\operatorname{MSE}(p,q)=\|p-q\|^2/d$.

Let $J_i=\partial v_i/\partial\theta$ and $F_i=a_i(u_i-v_i)+\omega_i(v_{{\rm a},i}-v_i)$. With sample coefficients held fixed, the parameter update direction is
\begin{equation}
 -\nabla_\theta\mathcal L
 =\frac1n\sum_iW_iJ_i^\top F_i
   -\lambda\nabla_\theta\mathcal R_\star.
\end{equation}
Using the stop-gradient target $y_i=\operatorname{sg}[v_i+F_i]$, the surrogate regression loss $\frac12W_i\|v_i-y_i\|^2$ produces the exact first-order parameter gradient $-W_iJ_i^\top F_i$ when evaluated at the current forward pass. Under the coordinate-mean squared error convention, the implemented matching loss $5t_i^2\operatorname{MSE}(v_i,y_i)$ corresponds to $W_i=10t_i^2/d$, and the frozen-reference penalty $10^{-4}\operatorname{MSE}(G_\theta,G_{\theta_0})$ corresponds to $\lambda=2\times10^{-4}/d$ in the half-squared-Euclidean-norm formulation.

\subsection{Guided-reference interpretation}

For a fixed prompt $c$, let $\pi_\theta^\gamma$ denote the endpoint distribution induced by the guided field $G_\theta^\gamma$, a specified solver and discretization, and an initial-noise law. Set $\mu_\star=\pi_{\theta_0}^{\gamma_\star}$ and $\mu_b=\pi_{\theta_0}^1$. Assuming positive densities on the support of $q$ with respect to a common dominating measure and finite log-ratio expectations,
\begin{equation}
 D_{\rm KL}(q\|\mu_\star)
 =D_{\rm KL}(q\|\mu_b)
   -\mathbb E_q\log\frac{\mu_\star}{\mu_b}.
 \label{eq:supp-reference-motivation}
\end{equation}
Thus changing the reference from $\mu_b$ to $\mu_\star$ in a KL-regularized reward objective adds the implicit reward $\tau\log(\mu_\star/\mu_b)$ for regularization strength $\tau>0$.

The implemented penalty $\mathcal R_\star$ instead matches guided velocities on forward-noised samples. It is not identified with the endpoint KL above. Writing $\Delta_c=v_\theta^c-v_{\theta_0}^c$ and $\Delta_u=v_\theta^\emptyset-v_{\theta_0}^\emptyset$, its residual is
\begin{equation}
 \Delta_G=\gamma_\star\Delta_c+(1-\gamma_\star)\Delta_u.
\end{equation}
Unlike CFG-free ($\gamma=1$) or split-branch penalties, it constrains the combination used during sampling. For $\gamma_\star\neq1$, changes $\Delta_c=h$ and $\Delta_u=\gamma_\star h/(\gamma_\star-1)$ cancel in this residual, so matching the guided field does not require matching each branch separately.

\section{Experimental Details}
\label{sec:supp-details}

\paragraph{Optimization and sampling hyperparameters.}
All methods fine-tune SD3.5-Medium using LoRA (rank $32$, $\alpha=64$) at $512\times512$ resolution in BF16 precision. We use AdamW with learning rate $3\times10^{-4}$, $(\beta_1,\beta_2)=(0.9,0.999)$, $\epsilon=10^{-8}$, weight decay $10^{-4}$, and gradient norm clipping at $1.0$. Each rollout batch comprises $48$ prompts with $24$ sampled images per prompt ($1{,}152$ images total), using a $10$-step DPM2 solver, $9$ subsampled training timesteps per image, $1$ inner epoch, and $1$ optimizer step per rollout batch. Evaluations use a $40$-step Euler solver at $512\times512$ resolution.

\paragraph{Method-specific loss details.}
AWM uses its per-timestep clipped likelihood-ratio objective with generalized-Huber power $0.25$, clipping radius $1.0$, and an EMA velocity-MSE penalty of weight $1.0$~\cite{xue2025advantage}. DiffusionNFT uses its adaptive positive/negative regression loss with $\beta_{\rm NFT}=0.1$ on OCR and $\beta_{\rm NFT}=1.0$ on GenEval and PickScore, normalizing positive and negative clean-latent residuals by separate detached mean-absolute-error (MAE) factors floored at $10^{-5}$~\cite{zheng2026diffusionnft}. RAM uses the detached regression target $v_{\rm ref}+cA(u-v_{\rm a})$ with unscaled clipped advantage $A$, $c=100$ on GenEval/OCR, $c=1000$ on PickScore, and uniform time weighting~\cite{bergmeister2026ram}. All methods include a frozen-reference penalty weight of $10^{-4}$ in coordinate-MSE units.

\FloatBarrier
\end{document}